\documentclass{article}

    \PassOptionsToPackage{numbers, compress}{natbib}

    \usepackage[preprint]{neurips_2019}

\usepackage{times}
\usepackage{graphicx} 
\usepackage{subfigure}
\usepackage{amsmath}
\usepackage{amsfonts}
\usepackage{color}
\usepackage{url}
\usepackage{wrapfig}

\usepackage{algorithm}
\usepackage{algorithmic}

\usepackage[utf8]{inputenc} 
\usepackage[T1]{fontenc}    
\usepackage{hyperref}       
\usepackage{url}            
\usepackage{booktabs}       
\usepackage{amsfonts}       
\usepackage{nicefrac}       
\usepackage{microtype}      
\usepackage{xcolor}

\newcommand{\lcom}[1]{}

\title{Learning an Adaptive Learning Rate Schedule}

\author{%
 Zhen Xu, Andrew M. Dai, Jonas Kemp, Luke Metz \\
 Google Research \\
 \texttt{zhenxu,adai,jonasbkemp,lmetz@google.com}
}

\begin{document}

\maketitle

\begin{abstract}
  The learning rate is one of the most important hyper-parameters for model training and generalization. However, current hand-designed parametric learning rate schedules offer limited flexibility and the predefined schedule may not match the training dynamics of high dimensional and non-convex optimization problems. In this paper, we propose a reinforcement learning based framework that can automatically learn an adaptive learning rate schedule by leveraging the information from past training histories. The learning rate dynamically changes based on the current training dynamics. To validate this framework, we conduct experiments with different neural network architectures on the Fashion MINIST and CIFAR10 datasets. Experimental results show that the auto-learned learning rate controller can achieve better test results.
  In addition, the trained controller network is generalizable -- able to be trained on one data set and transferred to new problems.
\end{abstract}

\vspace{-4mm}
\section{Introduction}
\vspace{-2mm}
The learning rate is often regarded as the single most important hyper-parameter to tune and highly influences model training using gradient decent algorithms \cite{BeYo12, GoYo16}. Researchers have developed several learning rate schedules such as linear decay, cosine decay, exponential decay, inverse square root decay, etc., sometimes with warm up steps, for different optimization problems \cite{ScZh13, ZeMa12}. However, there is limited intuition about which learning rate schedule best suits a given problem. In practice, researchers adopt a trial-and-error approach for different learning rate schedules along with different hyper-parameters, which is very time consuming \cite{BeBe12}.
In this paper, we would like to automatically learn a controller that adapts learning rate schedule by incorporating information from past training dynamics.
In addition, current learning rate schedules assume predefined parametric learning rate changes, which are fixed irregardless of actual training dynamics. The optimization landscape can be very complex \cite{LiZh18} and these parametric schedules have limited flexibility and may not be optimized for the training dynamics of different high dimensional and non-convex optimization problems. In comparison, our framework offers an auto-learned or meta-learned adaptive learning rate schedule that adapts dynamically based on current training dynamics. 

There are several related works proposing better update schedules for gradient descent algorithms. \cite{AnDe16} propose to directly learn the gradient descent updates using a long short-term memory (LSTM) network. Our work only learns the learning rate and so is more efficient. 
Hypergradient takes the derivative of the learning rate and updates the learning rate based on its gradient \cite{BaCo17}. In addition to the current state, our approach also considers the entire training history and has a more comprehensive view.
\cite{DaTa16} propose to use reinforcement learning (RL) to adapt the learning rate. 
In comparison, we use validation loss as the reward signal and a learning rate scaling function as the action. They improve the generalization capability and stability.
There are a family of widely used optimizers that dynamically adapt the learning rate on a per-parameter basis. For example, Adagrad adapts the learning rate per weight based on the sums of the squares of the gradients \cite{DuHa11}, while Adam uses an exponentially decayed average of past gradients \cite{KiBa15}. However, these optimizers still require a global learning rate which is important to tune. Our work is complementary to these works.

This paper makes three main contributions: First, we propose a reinforcement learning based framework to automatically learn an adaptive learning rate schedule based on past training histories. This schedule can adjust the learning rate dynamically to adapt to current training dynamics. Second, we present an effective set of state observation features, reward functions, and actions for the learning rate decision problem. Specifically, different from the previous work, we use validation loss as the reward signal and a learning rate scaling function as the action. Third, we conduct  experiments on Fashion MNIST and CIFAR10 datasets with convolutional neural networks (CNN) \cite{LeBo98} and residual networks (ResNet) \cite{HeZh16a, HeZh16b} to show the effectiveness and generalization capability of our framework. The auto-learned learning rate schedule can achieve better results and generalize to different datasets.



\vspace{-2mm}
\section{An Auto-learned Adaptive Learning Rate } \label{sec:framework}
\vspace{-2mm}

\subsection{Controller and Trainee Network}
\vspace{-1mm}
In our framework, we use RL to train a learning rate controller, which proposes learning rates using features from the training dynamics of the trainee network.
The trainee network is trained for a certain number of steps using a proposed learning rate, reports the observations of training dynamics to the controller which then returns a new learning rate. The whole process keeps running until reaching a certain stopping criterion. 

\subsection{State Observation, Reward, Action}
\vspace{-1mm}

\noindent \textbf{Observation} In order to characterize the training dynamics, we design a set of state observation features $\mathbf{s}$, which is an extension of the features used in \cite{DaTa16}. The proposed features include the current train loss, validation loss, variance of network predictions, variance of network prediction changes, mean and variance of the weight matrix of the final dense layer, and the previous step learning rate. As a general principle, we want our features to be easy to compute and generalizable to different trainee model architectures. This is why we only use the moments of the weight matrix of the final dense layer instead of all layers. 

\noindent \textbf{Reward} We use the per step validation loss as the reward $r$. Even though the final validation loss is what we really care about, empirically, we find that providing a reward $r_t$ for each training step can achieve better results compared with just using the final reward. 
These intermediate rewards provide more direct feedback and make credit assignment much easier.

\noindent \textbf{Action} The most direct action of the controller could be proposing a new learning rate. However, the learning rate is very sensitive and could be in the $10^{-6}$ scale or even smaller. It would be very unstable to directly use the network output as the learning rate. 
Another choice of action could be proposing the log of the learning rate. However, we want our action to be generalizable to different data sets since they may require different learning rate scales. 
Instead, we propose a learning rate scaling action $a_t$. At the first step, we provide a default learning rate. In the following steps, we use the network output as the scaling factor for the previous step learning rate, which can scaling it up or down. In this case, the controller can provide both warm up and decay capabilities in a stable way. 
This action provides a better inductive bias keeping learning consistent across steps.
 
\subsection{Auto-learned Adaptive Learning Rate Schedule with Proximal Policy Optimization}
\vspace{-1mm}
In this section, we present the RL-based learning rate controller, which is trained to propose a better learning rate schedule in order to reduce the validation loss in the model training process. 
Since the reward is validation loss and it is non-differentiable, we use policy gradients, and in particular the proximal policy optimization (PPO) algorithm \cite{ScWo17, ScLe15} to learn the controller parameters $\theta_c$ as it has better sample complexity. PPO optimizes a clipped surrogate objective function:
\begin{equation}
J(\theta_c) = E_t\Big[\min(w_t(\theta_c)A_t, \mathrm{clip}(w_t(\theta_c), 1-\epsilon, 1+\epsilon)A_t)\Big] \label{eq:clip_loss}
\end{equation}
to penalize large policy updates. $w_t(\theta_c)$ 
is the importance weighting probability ratio. $A_t$ is the advantage function at time $t$. $\epsilon$ is a hyper-parameter. 

\vspace{-2mm}
\section{Experiment Setting} \label{sec:experiment_setting}
\vspace{-2mm}

\subsection{Data Set and Model Architecture}
In our experiments, we use two data sets: Fashion MNIST \cite{XiRa07} and CIFAR-10 \cite{KrHi10}. 
For Fashion MNIST, we use 50k, 10k, and 10k images as the training, validation and test sets. For CIFAR10, we use 40k, 10k, and 10k images as the training, validation and test sets. 
We test on two model architectures: CNN and ResNet. CNN follows the structure of LeNet \cite{LeBo98}.
ResNet follows the structures from \cite{HeZh16a, HeZh16b}. We use the open source TensorFlow implementation \cite{Tens19}. 

\subsection{Baseline Learning Rate Schedule}
\vspace{-1mm}
We use a popular step decay learning rate schedule as our baseline \cite{GeKa19}, which is used in the open source ResNet implementation \cite{Tens19}. It contains three components: initial learning rate, discount step, and discount factor. The baseline schedule starts from the initial learning rate, then it decreases by the discount factor every discount steps. 
In the baseline experiments, we test all combinations from the initial learning rate in $[0.1, 0.01, 0.001, 0.0001]$, the discount step in $[10, 20, 50, 100]$, and the discount factor in $[0.99, 0.9, 0.88]$. After choosing the best baseline schedule, we run it 10 times with the same set of hyper-parameters and report mean and standard deviation of test loss and accuracy. 

\begin{table}[t]
\caption{Performance comparison between baseline and auto-learned learning rate schedules. We report the mean and standard deviation of the test set loss and accuracy derived from 10 runs with the same set of hyper-parameters. Parentheses denote standard deviations. We bold the best numbers for each model/dataset pair. Asterisk denotes the improvement is statistical significant under independent two-sample t-test with p-value threshold $0.05$. Fa. stands for Fashion.}
\label{table:experiment_result}
\resizebox{\columnwidth}{!}{
\centering
\begin{tabular}{cccccc}
\toprule
& & \multicolumn{2}{c}{\textbf{Baseline}} & \multicolumn{2}{c}{\textbf{Auto-learned}} \\
Dataset & Model & Test Loss & Test Accuracy & Test Loss & Test Accuracy \\
\midrule
Fa. MNIST & CNN    & 0.2497 (0.0042) & 0.9102 (0.0019) & \textbf{0.2351}$^{\ast}$ (0.0038) & \textbf{0.9201}$^{\ast}$ (0.0022)  \\
Fa. MNIST & ResNet & 0.2346 (0.0074) & 0.9188 (0.0029) & \textbf{0.2296} (0.0069) & \textbf{0.9192} (0.0028)  \\
\midrule
CIFAR10       & CNN    & 0.9539 (0.0140) & 0.6759 (0.0048) & \textbf{0.9361}$^{\ast}$ (0.0104) & \textbf{0.6787} (0.0041)  \\
CIFAR10       & ResNet & 0.8317 (0.0155) & 0.7395 (0.0206) & \textbf{0.6288}$^{\ast}$ (0.0196) & \textbf{0.8181}$^{\ast}$ (0.0069) \\
\bottomrule
\end{tabular}
}
\end{table}

\begin{figure}[t!]
\centering 
\subfigure[Fa. MNIST CNN] {\label{fig:mnist_cnn}\includegraphics[width=0.24\textwidth]{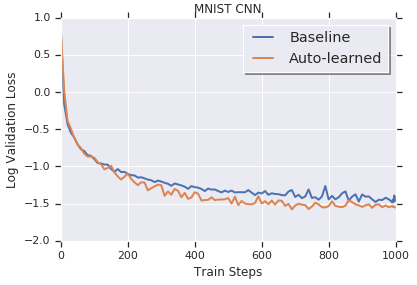}}
\subfigure[Fa. MNIST ResNet] {\label{fig:mnist_resnet}\includegraphics[width=0.24\textwidth]{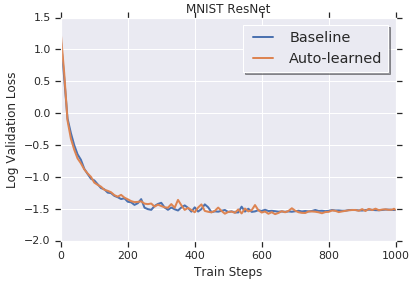}}
\subfigure[CIFAR10 CNN] {\label{fig:cifar10_cnn}\includegraphics[width=0.24\textwidth]{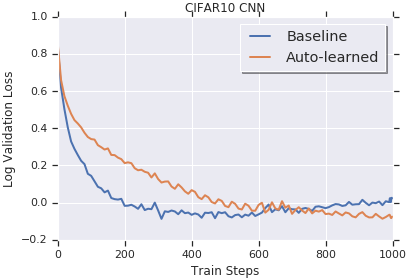}}
\subfigure[CIFAR10 ResNet] {\label{fig:cifar10_resnet}\includegraphics[width=0.24\textwidth]{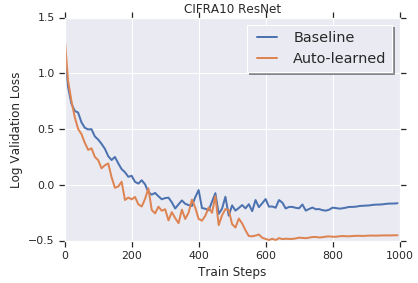}}
\protect\caption{Training trajectory comparison between baseline and auto-learned learning rate schedules. The $y$ axis is the log of validation loss. The $x$ axis is the number of training steps.
\lcom{style nit: increase font sizes on all figures. Remove gray background.}}
\label{fig:valid_loss_plot}
\end{figure}

\subsection{Training Details}
\vspace{-1mm}
We use validation loss as the reward signal and compare the hold-out test set loss and accuracy between baseline learning rate schedules and our auto-learned schedules. 
For trainee network training, the initial learning rate is in $[0.01, 0.001, 0.0001]$, the train batch size is 1k, and we run 1k training steps for all model architectures on all data sets.
Given 50k training set size, 50 train steps equal to 1 train epoch. In total, we train 20 epochs for Fashion MNIST and 25 epochs for CIFAR10. 
We will use the number of train steps for future plots.
In experiments, the controller proposes a new learning rate every 10 training steps. We refer to the whole 1k train steps of the trainee network as one training episode. The controller network is trained after every training episode of the trainee network. 
The actor of the controller is a multilayer perceptron (MLP) that contains one hidden layer with size 32. We also test a LSTM actor with hidden size 32. The critic is a MLP that contains one hidden layer with size 32. The learning rate is 0.001 for actor and 0.005 for critic.
Note that the main goal of our experiments is to show the effectiveness of the auto-learned learning rate schedule instead of outperforming the state-of-the-art classification accuracy on the target task. 
In our experiments, we restrict the training epoch to 25 and use ResNet with 18 total layers for computational reasons.

\vspace{-4mm}
\section{Experiment Results} \label{sec:experiment_results}
\vspace{-2mm}

\subsection{Test Set Results And Training Trajectories}
\vspace{-1mm}
Table \ref{table:experiment_result} shows the performance comparison between baseline and auto-learned learning rate schedules on test loss and accuracy. We choose the checkpoint based on the best validation loss and evaluate it on the test set once to get the single run test loss and accuracy.
From this table, we can see the auto-learned learning rate schedule achieves better results on all tasks. 
We hypothesize because it does not follow the predefined parametric learning rate changes and has the higher flexibility to adapt the learning rate based on the training dynamics.
Figure \ref{fig:valid_loss_plot} shows the training dynamics comparison between baseline and  auto-learned learning rate schedules in terms of the log of validation loss. 
For CNN on Fashion MNIST and ResNet on CIFAR10 in Figure \ref{fig:mnist_cnn} and Figure \ref{fig:cifar10_resnet}, the auto-learned schedule achieves lower validation loss faster. 

\begin{figure}[t!]
\centering 
\subfigure[Fa. MNIST CNN] {\label{fig:mnist_cnn_lr}\includegraphics[width=0.24\textwidth]{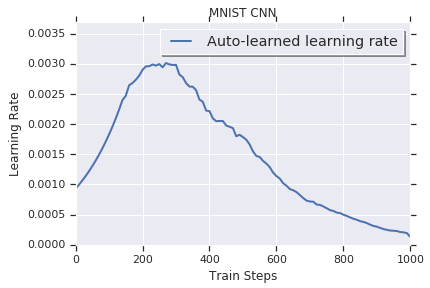}}
\subfigure[Fa. MNIST ResNet] {\label{fig:mnist_resnet_lr}\includegraphics[width=0.24\textwidth]{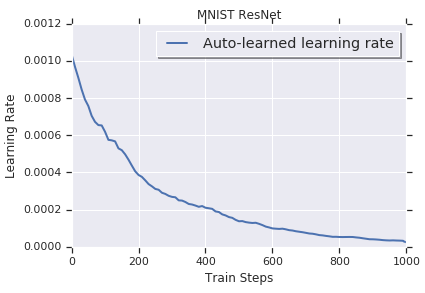}}
\subfigure[CIFAR10 CNN] {\label{fig:cifar10_cnn_lr}\includegraphics[width=0.24\textwidth]{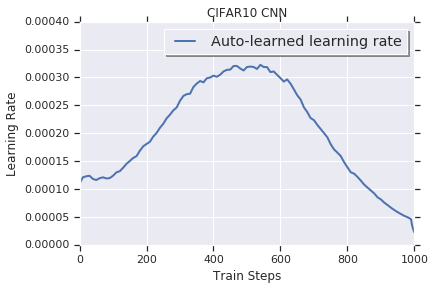}}
\subfigure[CIFAR10 ResNet] {\label{fig:cifar10_resnet_lr}\includegraphics[width=0.24\textwidth]{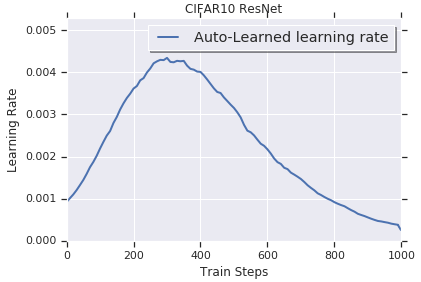}}
\protect\caption{Auto-learned learning rate schedule. The $x$ axis is the number of training steps. The $y$ axis is the learning rate. \lcom{If there is time it would be cool to put the baseline schedules also on these plots.}}
\label{fig:learning_rate_plot}
\end{figure}

\subsection{Auto-learned Learning Rate Schedule}
In this section, we present the auto-learned learning rate schedules as shown in Figure~\ref{fig:learning_rate_plot}. For the CNN model on Fashion MNIST in Figure~\ref{fig:mnist_cnn_lr} and ResNet model on CIFAR10 in Figure~\ref{fig:cifar10_resnet_lr}, the controller learns to warm up first and then decay the learning rate. This fits human's intuition when designing some existing learning rate schedules, but it is automatically learned and does not follow a predefined trajectory. 
For ResNet model on Fashion MNIST in Figure~\ref{fig:mnist_resnet_lr}, the auto-learned learning rate schedule is similar to an exponential decay learning rate schedule.
For CNN model on CIFAR10 in Figure~\ref{fig:cifar10_cnn_lr}, the auto-learned learning rate keeps flat at first, then warms up and decays later.

\vspace{-2mm}
\subsection{Transferability}
\vspace{-1mm}
We also investigate the transferability of a trained controller network when applied to different datasets. 
In the experiments, we load the controller checkpoints of CNN and ResNet models on CIFAR10, and let them propose the learning rate schedule for CNN and ResNet models on Fashion MNIST accordingly without training the learning rate controller. 
In comparison, we also apply the best baseline learning rate schedules of models on CIFAR10 to Fashion MNIST. 
From Table \ref{table:transfer} we can see the trained controller is transferable between two data sets. 
Our learning rate controller does not simply memorize, but is able to learn a transferable procedure to tune learning rate that generalizes.

\begin{table}[t]
\renewcommand{\arraystretch}{1.2}
\caption{Performance comparison between transferred baseline learning rate schedule and transferred controller network from CIFAR10 to Fashion MNIST. We report the mean and standard deviation of the test set loss and accuracy derived from 10 runs with the same set of hyper-parameters. Asterisk denotes the improvement is statistical significant under independent two-sample t-test with p-value threshold $0.05$.}
\centering
    \label{table:transfer}
    \begin{tabular}{ccccc}
    \hline
           &    \multicolumn{2}{c}{\textbf{Transferred Baseline}}  &   \multicolumn{2}{c}{\textbf{Transferred Controller Network}}     \\
     Model &     Test Loss & Test Accuracy & Test Loss & Test Accuracy  \\
    \hline
     CNN    & 0.2730 (0.0031) & 0.9021 (0.0013) & \textbf{0.2598}$^{\ast}$ (0.0071) & \textbf{0.9074}$^{\ast}$ (0.0030) \\
    \hline
     ResNet & 0.2443 (0.0040)& 0.9166 (0.0025) & \textbf{0.2315}$^{\ast}$ (0.0072) & \textbf{0.9212}$^{\ast}$ (0.0034) \\
    \hline
    \end{tabular}
\end{table}


\vspace{-2mm}
\section{Conclusion} \label{sec:conclusion}
\vspace{-2mm}
In this paper, we propose a reinforcement learning based framework which can auto-learn an adaptive learning rate schedule based on the information from past training histories. In order to achieve this goal, we introduce an effective set of features to characterize the dynamic training process, meaningful reward function, action space, and a sample-efficient RL algorithm to adapt the learning rate dynamically. Experimental results on Fashion MNIST and CIFAR10 data sets with CNN and ResNet models show our framework can learn a better learning rate schedule compared with step decay baseline schedules and can be transfered to new datasets never seen during meta-training.


\bibliography{main}
\bibliographystyle{unsrtnat}

\end{document}